%% file: main.tex
\documentclass{article}

\usepackage[final,eandd]{neurips_2026}

\usepackage[utf8]{inputenc}
\usepackage[T1]{fontenc}
\usepackage{hyperref}
\usepackage{url}
\usepackage{graphicx}
\usepackage{booktabs}       
\usepackage{array}
\usepackage{multirow}
\usepackage{amsmath}
\usepackage{amsfonts}       
\usepackage{amssymb}       
\usepackage{microtype}      
\usepackage[dvipsnames]{xcolor}       
\usepackage{algorithm}      
\usepackage{algpseudocode}  
\usepackage{tcolorbox}
\usepackage{colortbl}
\tcbuselibrary{breakable}

\newcommand{\datasetname}{JRDB-AVR}
\newcommand{\methodname}{\datasetname{}-Agent}
\newcommand{\baselineMonolithic}{{Monolithic VLM}}
\newcommand{\baselineCot}{{Chain-of-Thought}}
\newcommand{\baselineSearch}{{Search-Recognize-Pipeline}}
\newcommand{\baselineReact}{{ReAct}}
\definecolor{mygray}{gray}{.9}
\newtcolorbox[auto counter,number within=section]{prompt}[2][]{
    colback=white,
    colbacktitle=black!60,
    coltitle=white,
    fontupper=\footnotesize,
    boxsep=4pt,
    left=0pt,
    right=0pt,
    top=0pt,
    bottom=0pt,
    boxrule=0.8pt,
    breakable,
    title={Prompt \thetcbcounter: #2},
    #1,
}

\title{\datasetname{}: An Active Visual Reasoning Benchmark for Embodied Agents in Real-World Environments}

\author{%
  Zhixi Cai$^{*\dagger}$ \\
  Monash University \\
  \texttt{zhixi.cai@monash.edu}
  \And
  Fucai Ke$^{*}$ \\
  Monash University \\
  \texttt{fucai.ke1@monash.edu}
  \And
  Sukai Huang \\
  Monash University \\
  \texttt{sukai.huang@monash.edu}
  \AND
  Maria Garcia de la Banda \\
  Monash University \\
  \texttt{maria.garciadelabanda@monash.edu}
  \And
  Peter J. Stuckey \\
  Monash University \\
  \texttt{peter.stuckey@monash.edu}
  \AND
  Gholamreza Haffari \\
  Monash University \\
  \texttt{gholamreza.haffari@monash.edu}
  \And
  Hamid Rezatofighi \\
  Monash University \\
  \texttt{hamid.rezatofighi@monash.edu}
}

\begin{document}

\maketitle

\begingroup
\renewcommand{\thefootnote}{\fnsymbol{footnote}}
\footnotetext[1]{Equal contribution.}
\footnotetext[2]{Corresponding author.}
\endgroup

\begin{abstract}
In complex embodied visual reasoning scenarios, an agent often has only a limited field of view, and the evidence needed to answer a question may be distributed across time, viewpoint, and interacting objects. A model may therefore give a plausible answer without ever observing the relevant object, time, or view that supports it. Current visual reasoning benchmarks largely evaluate passive observations and final answers, overlooking settings that require active reasoning and evidence acquisition. We introduce \datasetname{}, a benchmark derived from existing real-world JRDB robotics data through a structured question-generation engine that turns this gap into an explicit evaluation: an embodied agentic system receives a visual reasoning question, requests bounded observations by timestamp and viewing angle, and is evaluated on both the final answer and the grounded visual evidence supporting it. The benchmark contains diverse questions over multiple real-world environments involving temporal search, viewpoint selection, and human-oriented compositional reasoning. We also introduce \methodname{}, a reference active reasoning agentic method that maintains an explicit observation-grounded graph-based world model and answers through solving. Experiments reveal a substantial gap between answer accuracy and evidence accuracy in current baselines, showing that current VLMs can produce unsupported correct answers and that active evidence-aware evaluation is necessary for embodied visual reasoning. Code and benchmark are available at \url{https://github.com/ControlNet/JRDB-AVR}.
\end{abstract}

\begin{figure}[t]
\centering
\includegraphics[width=\linewidth]{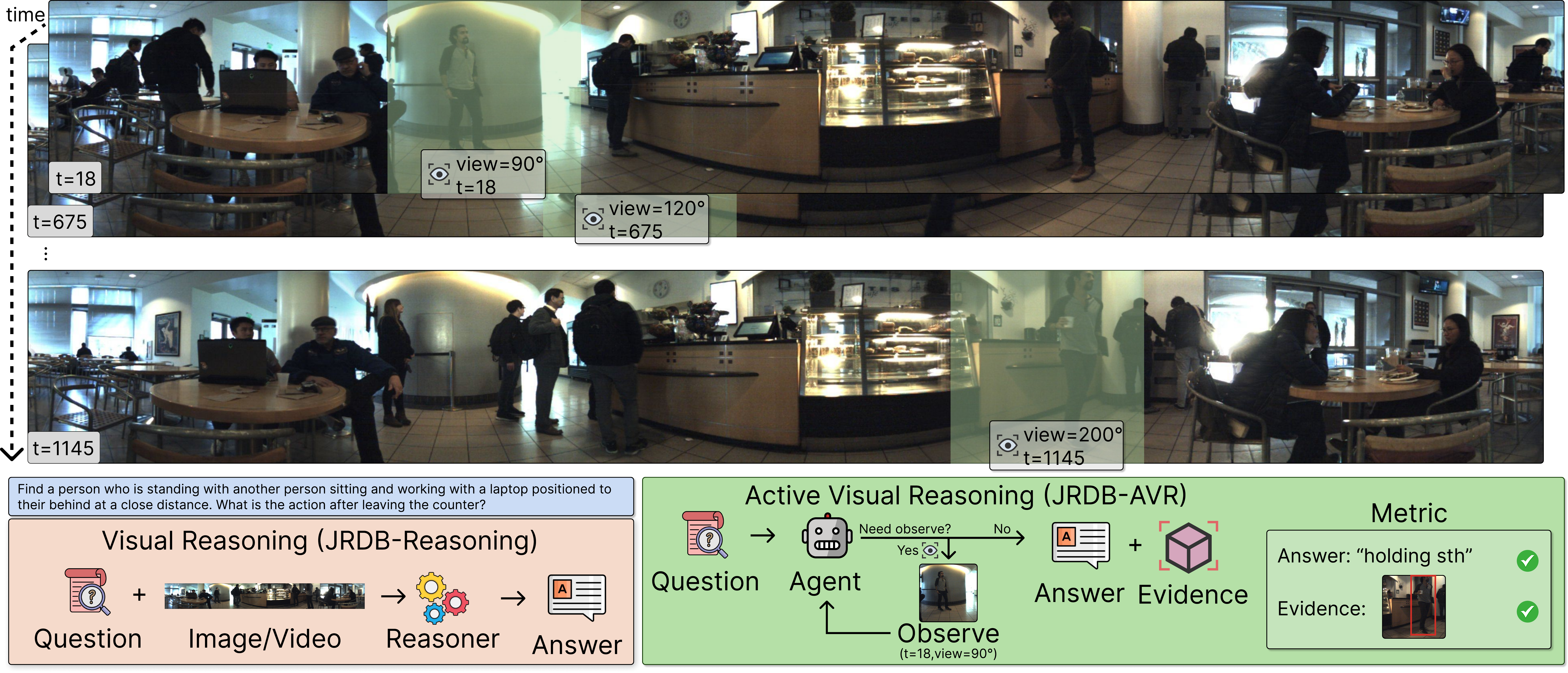}
\caption{Overview of \datasetname{}. Unlike passive visual reasoning, where relevant image/video observations are pre-selected and only the final answer is evaluated, \datasetname{} requires an agent to actively observe across time and view angle, then evaluates both the answer and visual evidence.}
\label{fig:teaser}
\end{figure}

\section{Introduction}
\label{sec:introduction}
Embodied agents often need to answer questions in environments that they cannot fully observe at once. Since a robot camera has a limited field of view, the evidence needed for reasoning may lie in another direction or appear at another time. Consider a robot in a crowded cafe asked whether the person near the counter was carrying a bag before joining a group. The question cannot be reduced to simply recognizing an object in one image: the required evidence may appear in an earlier frame, a different viewing angle, or a particular person among several visually similar people. In this setting, answer correctness alone is an unsafe signal for visual reasoning capability. A system can guess the right answer from priors or partial context while grounding it in the wrong person, time, or view, which is exactly the hallucination an embodied agent must avoid.

These failure cases reveal a broader evaluation gap. In embodied visual reasoning, success depends not only on producing a correct answer, but on deciding where to look, when to look, and whether the acquired observation actually supports the answer. Existing video reasoning benchmarks evaluate passive-input capabilities~\citep{hudsonGQA2019,yiCLEVRER2019,wuSTAR2021,leiTVQA2018, leiTVQA2020}, and embodied human-scene datasets such as JRDB-Reasoning provide rich reasoning annotations over real-world robotic scenes~\citep{jahangardJRDBSocial2024,jahangardJRDBReasoning2026}. However, these settings still largely evaluate reasoning with passive perception, and typically score the final answer rather than the supporting evidence. This setup can therefore overstate embodied visual reasoning ability, especially in complex real-world environments where an agent must resolve temporal ambiguity, choose the right viewpoint, and identify the correct person in a crowded scene.

This gap has become sharper as recent VLMs and tool-using agents produce increasingly accurate answers and multi-step trajectories~\citep{liBLIP22023,zhuMiniGPT42023,baiQwen25VL2025,yaoReAct2023,surisViperGPT2023,gaoCLOVA2024}. A correct answer or detailed reasoning trace is still not evidence that the agent inspected the right objects. Recent VLMs can answer plausibly from language priors, dataset bias, or partial visual context even when the relevant visual evidence is missing or incorrectly localized~\citep{gouEmpirical2025}. If evaluation fixes the observation in advance or scores only the final answer, unsupported success remains hard to distinguish from grounded reasoning. For active embodied systems, this is part of the task definition instead of a minor interpretability issue.

We therefore introduce \datasetname{}, a benchmark for real-world embodied active visual reasoning derived from the existing JRDB dataset~\citep{martin-martinJRDB2023}. Creating real active embodied reasoning data is difficult because a live robot cannot simultaneously observe every direction and every moment. \datasetname{} addresses this by repurposing JRDB panoramic robot videos into an observation interface: an agent receives a visual reasoning question and can request observations by timestamp and viewing angle. Instead of curating static fixed-view VQA examples, we construct \datasetname{} using a refined question-generation pipeline that derives active reasoning questions from JRDB annotations~\citep{jahangardJRDBSocial2024,jahangardJRDBReasoning2026,ehsanpourJRDBAct2022,leJRDBPanoTrack2024,saadatnejadJRDBTraj2023,biswasJRDBPose3D2026,vendrowJRDBPose2023}. \datasetname{} formalizes active observation as the benchmark protocol, and is evaluated on both the final answer and the grounded bounding box as visual evidence. The benchmark focuses on multi-step active visual reasoning over real-world crowded scenes, including temporal search, viewpoint selection, and compositional multi-object reasoning. Figure~\ref{fig:teaser} illustrates this shift in evaluation: rather than judging a model only by whether it can produce a plausible answer from partial context, \datasetname{} asks whether the agent actively acquires the right observation and grounds its answer in visual evidence.

To study how current VLM and agentic baselines perform under this benchmark, we introduce \methodname{} as a reference baseline for active reasoning on \datasetname{}. \methodname{} converts each question into a structured graph plan, actively observes the scene, writes observation-grounded entities, attributes, and relations into an explicit world model, and answers by solving the resulting graph state. Current results show that answer performance and evidence performance can diverge substantially: models may obtain correct answers without providing correct visual support. Thus, answer-only evaluation obscures important embodied reasoning failures.

The contributions of this paper are:
\begin{enumerate}
    \item We introduce \datasetname{}, a benchmark for real-world embodied active visual reasoning, together with a benchmark generation engine that derives active reasoning questions from JRDB annotations. We also define an evaluation protocol and metrics that report answer, evidence, and combined correctness, making the answer-evidence gap explicit when correct answers rely on unsupported visual evidence.
    \item We provide \methodname{}, a reference active reasoning baseline that uses a structured graph plan, an observation-grounded world model, and a graph solving method to produce answer-evidence predictions.
    \item We present an important empirical finding from the current evaluation protocol: strong baselines can reach materially different answer and evidence performance, showing that active visual reasoning needs evidence-aware evaluation rather than answer accuracy alone.
\end{enumerate}

\begin{table}[t]
\centering
\caption{
Comparison with related visual reasoning benchmarks.
We compare whether each benchmark requires temporal reasoning, spatial or viewpoint reasoning, the data domain and camera setting, whether observations can be actively selected, the form of localized visual evidence, and the final evaluation target.
\emph{Time + View} denotes active selection of both timestamp and viewing angle, while \emph{Time + BBox} denotes evidence localized by timestamp and target bounding box.
}
\label{tab:benchmark-comparison}
\small
\resizebox{\linewidth}{!}{
\begin{tabular}{lccccccc}
\toprule
Benchmark &
Temporal &
Spatial/View &
Domain &
Camera &
Active Obs. &
Evidence &
Evaluation \\
\midrule
GQA~\citep{hudsonGQA2019}
& $\times$ & \checkmark & Real & Static & $\times$ & $\times$ & Answer \\

CLEVRER~\citep{yiCLEVRER2019}
& \checkmark & $\times$ & Sim & Static & $\times$ & $\times$ & Answer \\

STAR~\citep{wuSTAR2021}
& \checkmark & \checkmark & Real & Static & $\times$ & $\times$ & Answer \\

MindCube~\citep{wang2025mindcube}
& $\times$ & \checkmark & Real & Static & $\times$ & $\times$ & Answer \\

VIEW2SPACE~\citep{keVIEW2SPACE2026}
& $\times$ & \checkmark & Sim & Static & $\times$ & BBox & Answer \\

JRDB-Reasoning~\citep{jahangardJRDBReasoning2026}
& \checkmark & \checkmark & Real & Moving & $\times$ & $\times$ & Answer \\

\textbf{\datasetname{}}
& \checkmark
& \checkmark
& \textbf{Real}
& \textbf{Moving}
& \textbf{Time + View}
& \textbf{Time + BBox}
& \textbf{Answer + Evidence} \\
\bottomrule
\end{tabular}}
\end{table}

\section{Related Works}
\label{sec:related-work}
\paragraph{Visual reasoning benchmarks.}
Previous visual reasoning benchmarks have established protocols for testing compositional questions, temporal events, and evidence in fixed visual inputs~\citep{ma2026iv}. For example, GQA emphasizes compositional question answering over images~\citep{hudsonGQA2019}, CLEVRER tests causal and temporal reasoning in synthetic videos~\citep{yiCLEVRER2019}, STAR targets situated video question answering~\citep{wuSTAR2021}, MindCube studies spatial mental modeling from limited views~\citep{wang2025mindcube}, and TVQA+ links questions to spatio-temporal evidence in video~\citep{leiTVQA2020}. Recent grounded video QA evaluation also asks whether a correct answer is visually supported by the relevant evidence~\citep{xiaoCan2024, ma2026iv, keVIEW2SPACE2026,du2026weatherreasonseg}. Temporal grounding and spatio-temporal video understanding benchmarks further study when events occur and where relevant evidence lies across clips~\citep{shouTemporal2016,buchEndtoend2019,wangVideoTree2024,yangUnleashing2025}. JRDB-based datasets bring this reasoning into crowded embodied scenes with social structure and robot-centric sensing~\citep{jahangardJRDBSocial2024,jahangardJRDBReasoning2026}. However, existing benchmarks generally evaluate either reasoning over fixed visual inputs or grounding within already provided videos. As summarized in Table~\ref{tab:benchmark-comparison}, \datasetname{} changes the benchmark protocol: an agent must actively acquire observations in a real-world embodied environment, and the final prediction is evaluated by both answer correctness and whether the acquired evidence supports the answer.

\paragraph{Visual reasoning methods.}
A common starting point for visual reasoning~\citep{keExplain2025} is to use general-purpose VLMs as monolithic predictors over fixed image or video inputs~\citep{liBLIP22023,zhuMiniGPT42023,chenInternVL2024,wangQwen2VL2024,baiQwen25VL2025}. To improve interpretability and control, early compositional methods explicitly decomposed questions into modular networks or executable programs~\citep{johnsonInferring2017}. More recent compositional pipelines expose intermediate programs, localized zoom-and-refinement stages, grounded actions, or tool calls~\citep{tiongPlugandPlay2022,surisViperGPT2023,yuZoomRefine2025,shenHuggingGPT2023,huang2026wetalkabout,luChameleon2023,guptaVisual2023}. Agentic and tool-integrated methods then extend this pattern toward multi-step reasoning with iterative action or tool use, feedback, memory, or explicit state~\citep{yaoReAct2023,gaoCLOVA2024,gouToRA2024,keHYDRA2024,keDWIM2025,wuVToolR12025,chenMemoryAnchored2025,caiNAVER2025,caiMATA2025,maDrVideo2024,chen2026hypospace}, with related work also studying LLM-symbolic planning and Bayesian, hierarchical, and active goal recognition under uncertainty~\citep{huang2025planning,zhangHuman2024,zhang2025probabilistic,zhangprobabilistic2026,zhang2026neurosymbolic}. In parallel, neuro-symbolic and structured-world-model approaches use explicit intermediate representations of entities and relations to support multi-step reasoning~\citep{yiNeuralSymbolic2018,kamaliNeSyCoCo2025,caiNEUSIS2025,huang2026minigran,li2026cotjudger}. These advances help motivate active visual reasoning, but they are usually evaluated on passive visual inputs. In contrast, \datasetname{} makes active evidence acquisition part of the benchmark requirement. To instantiate this protocol, we provide \methodname{} as a reference baseline with question-to-graph planning, observation-grounded world model maintenance, and graph solving.

\section{\datasetname{} Benchmark}
\label{sec:benchmark}
\datasetname{} evaluates active visual reasoning in real-world embodied environments. Each task instance asks a system to answer a question while deciding which observations are needed, and then to return both the answer and its visual evidence. The benchmark is designed so that the correctness of the answer and of the evidence can diverge: a system may guess the right answer from priors or partial context, but still fail if the evidence person, frame, viewpoint, or box does not support that answer.

\subsection{Dataset Generation}

\begin{figure}[t]
\centering
\includegraphics[width=\linewidth]{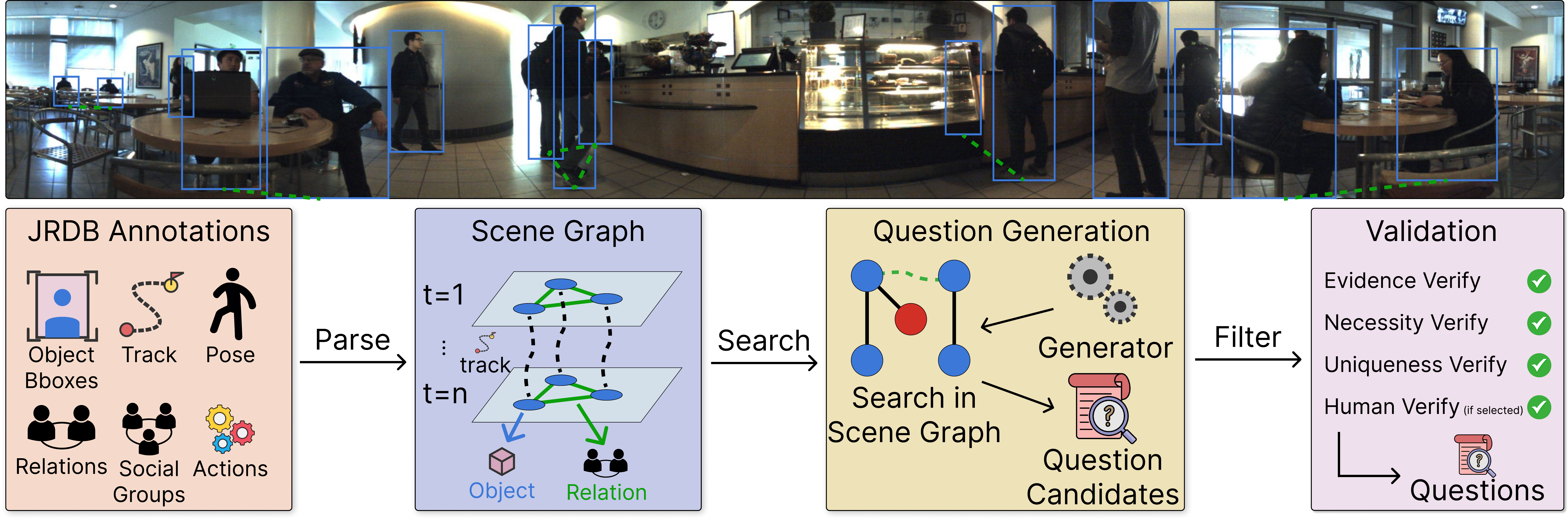}
\caption{Dataset generation pipeline for \datasetname{}. Starting from JRDB panoramic videos and annotations, we organize object boxes, tracks, poses, relations, social groups, actions, etc into temporal scene graphs. Question generators search these scene graphs for active reasoning patterns and produce candidate questions, which are retained only after they pass evidence, necessity, uniqueness, and manual quality checks.}
\label{fig:dataset-generation-pipeline}
\end{figure}

Figure~\ref{fig:dataset-generation-pipeline} summarizes the data generation process from the raw source data to \datasetname{}.

\paragraph{Source Data and Embodied Environments.}
\datasetname{} builds on the JRDB video dataset~\citep{martin-martinJRDB2023}, which includes crowded indoor and outdoor environments, person tracks, actions, spatial relations, and social context annotations~\citep{vendrowJRDBPose2023, ehsanpourJRDBAct2022, leJRDBPanoTrack2024, jahangardJRDBSocial2024, jahangardJRDBReasoning2026}. The benchmark is generated from complementary annotations provided by JRDB and its extension datasets. These include person identities, tracks, and bounding boxes from JRDB~\citep{martin-martinJRDB2023}; spatio-temporal actions from JRDB-Act~\citep{ehsanpourJRDBAct2022}; human poses from JRDB-Pose; panoptic tracks from JRDB-PanoTrack~\citep{leJRDBPanoTrack2024}; social-group and interaction annotations from JRDB-Social~\citep{jahangardJRDBSocial2024}; and human-object and geometric relations from JRDB-Reasoning~\citep{jahangardJRDBReasoning2026}.

\paragraph{Scene Graph Generation.}
We first generate a scene graph from the JRDB annotations for each sequence (video in JRDB dataset) that supports controlled sub-graph search for question generation. This graph organizes tracked objects, relations, attributes, and timestamps together with the facts needed for question generation, including person attributes, actions over time, robot-relative geometry, person-to-person relations, person presence, and bounding boxes, from which viewpoint-specific evidence is derived. In this way, dataset generation becomes a subgraph-search process rather than free-form question generation.

\paragraph{Question Candidate Generation.}
We then generate question candidates by querying this scene graph for patterns that induce active reasoning. The proposed benchmark is built from seven VQA generators, summarized in Table~\ref{tab:generator-families}. These generators create the main forms of active evidence search discussed in this work: localizing action temporal segments, following multi-hop relational paths, searching across time for a later state, and recovering a target from a unique anchor across time. Each question candidate is generated together with the benchmark fields needed for later evaluation, including the bounding boxes of evidence objects. Please refer to supplementary material for more details regarding question generators.

\begin{table*}[t]
\centering
\small
\caption{For each of the seven generators currently present in  \datasetname{}, the table provides a brief description of the generator, its question type and the metrics protocol of each question type.}
\label{tab:generator-families}
\begin{tabular}{>{\raggedright\arraybackslash}p{0.14\linewidth} >{\raggedright\arraybackslash}p{0.33\linewidth} >{\raggedright\arraybackslash}p{0.14\linewidth} >{\raggedright\arraybackslash}p{0.29\linewidth}}
\toprule[0.4mm]
\rowcolor{mygray} Name & Description & Type & Evaluation Protocol \\
\hline
Chain 2-hop & Ask about the target with 2-hop entity relations. & \multirow[t]{5}{0.14\textwidth}{Multiple choice} & \multirow[t]{5}{0.29\textwidth}{Correct if the selected choice matches the ground truth.} \\
Chain 3-hop & Extend the relational path to three hops before querying the target. &  &  \\
Chain fork-join & Merge two relational branches from a shared entity before querying the target. &  &  \\
Unique anchor & Locate a target from conditions and ask about a later state or action in another temporal location. &  &  \\
Long range & Track temporally distant consequences or long-range dependencies. &  &  \\
\hline
Chain hybrid & Combine a spatial relation with temporal and viewpoint search to locate the target timestamp and view angle. & Timestamp and viewpoint & Correct if the predicted timestamp is within $\pm 1$ second of the ground truth and the wrapped viewpoint error is within $\pm 20^\circ$. \\
\hline
Action boundary & Localize the timestamp at which an action begins. & Timestamp identification & Correct if the predicted frame is within $\pm 1$ second of the ground truth. \\
Action boundary & Localize the full temporal segment of an action. & Temporal localization & Correct if the predicted temporal segment has IoU $\ge 0.5$ with the ground truth. \\
\hline
Long range & Predict a temporally defined numeric gap or count over a longer-range relationship. & Numeric value & Correct if the prediction-to-ground-truth ratio lies in $[0.5, 2.0]$. \\
\bottomrule[0.4mm]
\end{tabular}
\end{table*}

\paragraph{Question Candidate Validation and Quality Assurance.}
Question candidates are not kept simply because a subgraph is matched. To ensure data quality, after generation, we automatically re-derive the stored answer from trusted JRDB annotations and retain only questions whose supporting evidence can be objectively recovered. This validation pass enforces evidence verifiability together with temporal necessity, multi-step necessity, view necessity, anchor uniqueness for unique-anchor rows, and hop necessity for chain questions. As a result, questions are removed when they remain answerable after perturbing the target time or collapsing an intermediate hop, as well as when their evidence cannot be evaluated reliably. Furthermore, the authors manually inspect a random subset of generated questions for verification.

\subsection{Dataset Statistics}

\datasetname{} is generated from the 27 JRDB test sequences using the corresponding JRDB and extension-dataset annotations. The released benchmark contains 2,098 active VQA questions. These counts describe the current released benchmark and should not be read as an upper bound on all valid questions derivable from the source annotations. The generation engine can be used to derive additional active visual reasoning questions from the same source data.

\subsection{Benchmark Protocol}
\label{sec:benchmark_protocol}

\paragraph{Task Definition.}
A task session contains a question $q$ and related metadata. The system receives $q$ and may request observations before producing a final answer. A completed prediction contains the answer $\hat{y}$ in the required format together with evidence $\hat{e}$, represented as the target localization by timestamp and bounding box. The task is active because the system is not given one fixed complete view up front, and it must choose where and when to look before answering.

\paragraph{Observation Interface.}
The observation interface as a tool is \texttt{observe(sequence, frame\_index, angle\_deg)} ($o = O(s,f,\theta)$). It returns a 480$\times$480 RGB cropped frame with some metadata rendered from stitched raw RGB JRDB videos (3760$\times$480, 15FPS). However, typically the agent does not receive the full panorama as the initial input. This choice keeps the benchmark close to embodied environments: the embodied agent system actively decides what to observe at a selected time and angle rather than reading the full scene at once.

\paragraph{Evaluation Metrics.}
The metrics separate what the agent answers from the answer's visual support. For each evaluated question $q_i$, we write $s_{\mathrm{ans}}^{(i)} \in \{0,1\}$ for answer correctness and $s_{\mathrm{evid}}^{(i)} \in \{0,1\}$ for evidence correctness. The answer score is a traditional VQA correctness score: multiple choice answers are checked through their choice index, frame answers through frame tolerance, temporal localization answers through IoU thresholding, numeric answers through ratio tolerance, and frame viewpoint answers through frame with angle tolerance. The evidence score checks whether the evidence matches the target support, requiring it to identify the correct target and, when box annotations are available, to produce a bounding box whose IoU exceeds a threshold in any frame where the target appears. We report the mean answer, evidence, and combined scores over $N$ evaluated questions:
\begin{equation}
\bar{s}_{\mathrm{ans}} = \frac{1}{N}\sum_{i=1}^{N} s_{\mathrm{ans}}^{(i)}, \qquad
\bar{s}_{\mathrm{evid}} = \frac{1}{N}\sum_{i=1}^{N} s_{\mathrm{evid}}^{(i)}, \qquad
\bar{s}_{\mathrm{comb}} = \frac{1}{N}\sum_{i=1}^{N} s_{\mathrm{ans}}^{(i)} s_{\mathrm{evid}}^{(i)}.
\label{eq:benchmark-metrics}
\end{equation}
The combined score therefore counts a question as correct only when both the answer and the evidence are correct for that question.
For the current benchmark, each question is constructed around a uniquely grounded target entity. Reasoning may require multiple intermediate entities and observations, but the final visual evidence is the target bounding box.

\section{\methodname{}}
\label{sec:agent}

\methodname{} is a training-free reference active reasoning baseline for \datasetname{}. It instantiates the benchmark's core challenge with an explicit world model: the agent must first observe the scene, write observation-grounded facts into a world model as memory, and then answer from the current world-model state. Because the relevant evidence may be scattered across time, viewpoint, and objects, a single observation or free-form reasoning trajectory is insufficient. The agent therefore maintains an explicit graph-based world model to accumulate observed entities, attributes, relations, viewpoints, and unresolved evidence  across interaction steps. A graph plan state guides the future observation and supports the final answer-evidence prediction. Together, these components make \methodname{} an evidence-aware active reasoning reference baseline.

\paragraph{Graph-query planning.}
Given a question, the agent first constructs a structured graph plan that specifies what must be grounded before answering. The plan contains a root node, a target node, candidate nodes, directed edges, and an answer specification. This converts the question from an unconstrained natural-language prompt into a graph query over entities and relations: the root node defines the starting evidence anchor, the target node defines the entity or event to recover, the edges define the relational or temporal path to follow, and the answer specification defines how the final answer should be read from the grounded graph. The plan is schema-validated before interaction, so later observations and world model maintenance are organized around an explicit reasoning target rather than an unconstrained chain of thought.

\paragraph{Observation-grounded world model.}
The world model in \methodname{} is an explicit graph memory rather than a latent or purely textual state. At step $t$, the world model $W_t$ stores entities, attributes, relations, and observation provenance. Entity records represent grounded person or object hypotheses, attribute records store local visual facts such as appearance, action, or state, relation records store spatial, temporal, or person-to-person relations, and observation records link these graph facts to the frame and viewpoint from which they were obtained. This use follows memory and neuro-symbolic views of world models as structured environment representations~\citep{haoReasoning2023a, caiNEUSIS2025}, but it is not a predictive future-frame or video-generation model. Its purpose is to make the agent's intermediate visual state readable, writable, and auditable.

\paragraph{Tool-based active world model maintenance.}
The agent interacts with the benchmark through a set of tools. It may call \texttt{get\_graph} to read the current world model state, \texttt{observe} to request a visual observation, \texttt{add\_entity} to add a grounded entity, \texttt{add\_attribute} to attach an observed property to an entity, \texttt{add\_relation} to record a grounded relation, and \texttt{final\_answer} to submit the final prediction. The observation action is \texttt{observe(sequence, frame\_index, angle\_deg)}, which returns a 480$\times$480 crop with metadata. Importantly, the changes of the world model are observation-grounded: an entity, attribute, or relation can be added only when it is tied to an already obtained frame-view observation. This interaction design lets the agent decide at each step whether to request another observation, update grounded information into the world model, or stop and produce a final answer.

\paragraph{Active reasoning loop.}
At inference, the agent receives only the visible question $q$ and sequence id $s$. It first builds a structured graph plan $P=\operatorname{Plan}(q)$ that specifies the anchor, intermediate, and target entities and initializes the world model $W_0=\operatorname{InitializeGraph}(P)$. At step $t$, the VLM policy of the agent $\pi$ selects a tool action:
\[
    u_t \sim \pi(P,W_t,q),
\]
where $u_t$ may be an observation request, a graph read, a graph write, or a final-answer action. The interaction is therefore a tool-based transition:
\begin{equation}
    W_{t+1} =
    \operatorname{ToolStep}(W_t, u_t; P, q, s),
    \qquad
    (\hat{y},\hat{e}) =
    \operatorname{Solve}(P,W_{\tau},q),
    \label{eq:agent-update}
\end{equation}
where $\tau$ is the stopping step, determined either by a \texttt{final\_answer} call or by the tool-call budget. For an observation action, \texttt{ToolStep} calls the public observation operator
$O(s,f,\theta)$ and registers the returned crop and metadata in $W_t$. For a graph-writing action, it applies \texttt{add\_entity}, \texttt{add\_attribute}, or \texttt{add\_relation} only if the proposed write is grounded in an already observed frame-view pair. Otherwise, the write is rejected and the graph state is unchanged. Thus, the loop alternates between acquiring observations and writing observation-grounded graph facts until the current graph state is sufficient for solving or the budget is exhausted.

\begin{figure}[t]
    \centering
    \includegraphics[width=\linewidth]{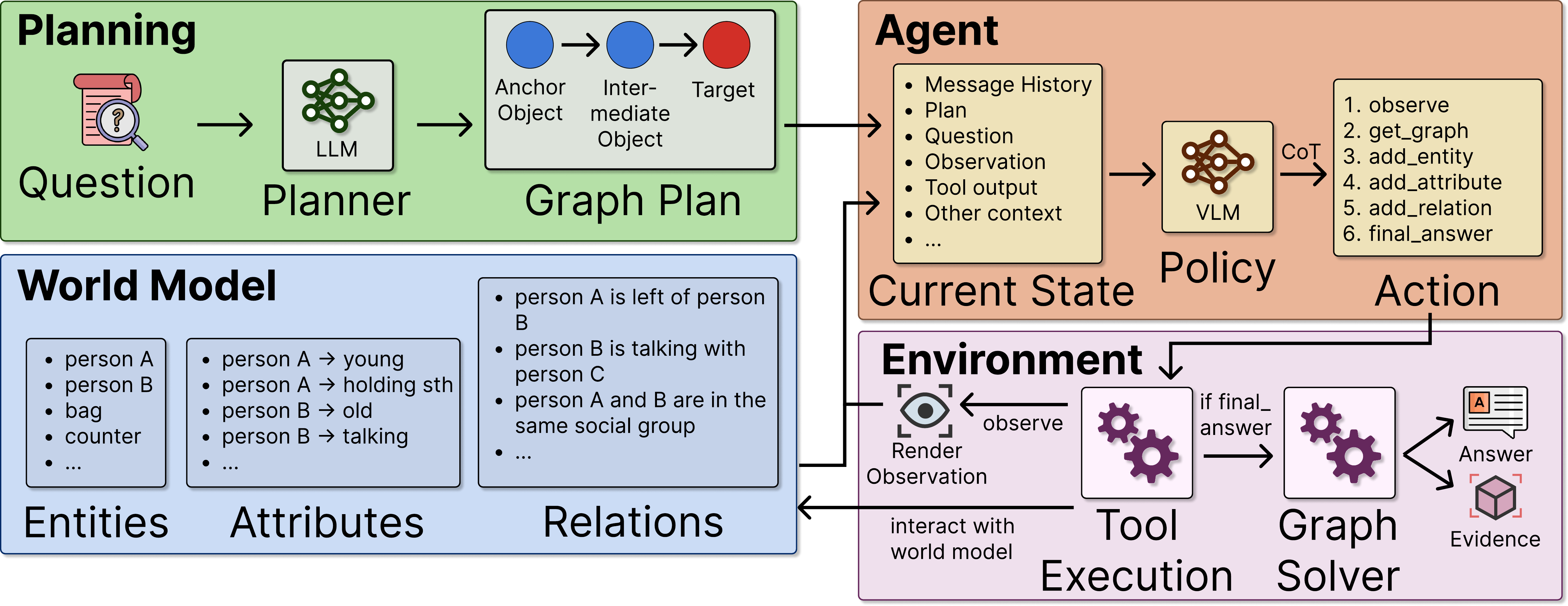}
    \caption{
    Pipeline of \methodname{}. The question is first converted into a structured graph plan that specifies the anchor, intermediate, and target entities. During interaction, the agent selects tool actions from the current state, requests observations when needed, and writes observation-grounded entities, attributes, and relations into an explicit world model. The final answer is produced by solving the graph state and returning both the answer and supporting visual evidence.
    }
    \label{fig:agent-pipeline}
\end{figure}

\paragraph{Solving and evidence-grounded output.}
When \texttt{final\_answer} is called, the agent does not simply accept a free-form answer. It first invokes a graph solver to solve the graph plan against the current world model. Concretely, the graph plan acts as a structured query, and the solver searches over graph facts accumulated from obtained observations to find a satisfying target. The proposed answer is then checked against this target, so that it is legal only if it is supported by the current graph state. The final output contains the predicted answer $\hat{y}$ and the supporting evidence $\hat{e}$.

\section{Experiments}
\label{sec:experiments}

\paragraph{Experimental setup.}
We evaluate on the released \datasetname{} benchmark, which contains 2,098 active VQA questions from 27 JRDB test sequences. All methods are evaluated on the same question ids and under the same evaluation protocol introduced in Section~\ref{sec:benchmark}. We report three primary metrics: the answer score $\bar{s}_{\mathrm{ans}}$, evidence score $\bar{s}_{\mathrm{evid}}$, and combined score $\bar{s}_{\mathrm{comb}}$ defined in Section~\ref{sec:benchmark_protocol}.

\paragraph{Baselines.}
We compare \methodname{} against the following four baseline methods which represent different levels of passive and active reasoning. \emph{\baselineMonolithic{}} answers from uniformly sampled panoramic context without active observation. \emph{\baselineCot{}}~\citep{weiChainofThought2022} uses the same visual input but adds intermediate reasoning before producing the answer and evidence. \emph{\baselineSearch{}} is a training-free localize-first workflow-based baseline inspired by zoom-and-refinement methods~\citep{yuZoomRefine2025}: it proposes a candidate target, requests an observation to refine the localization, and  answers from the refined evidence. \emph{\baselineReact{}}~\citep{yaoReAct2023} uses a tool-use agentic loop with the observation tool. Since \baselineMonolithic{} and \baselineCot{} cannot actively call tools for active observation, we give them a sampled panoramic context as passive input. This lets them access a broader pre-selected visual context, but they cannot decide where/when to observe.

\paragraph{Implementation details.}
All baseline methods are evaluated across six VLM backbones~\citep{googledeepmindGemma2026, baiQwen3VL2025, qwenteamQwen352026}: Gemma-4-E2B, Gemma-4-E4B, Qwen3-VL 4B, Qwen3-VL 8B, Qwen3.5 4B, and Qwen3.5 9B. All methods use the same benchmark questions, the same \texttt{observe(sequence, frame\_index, angle\_deg)} interface when observations are requested, and the same answer-evidence output schema. All methods are evaluated zero-shot on \datasetname{}. \methodname{} is evaluated on the same benchmark using the active reasoning design described in Section~\ref{sec:agent}, including its explicit graph-based world model, and uses Qwen3.5 9B, which achieves the strongest baseline combined score under \baselineReact{}. All reported experiments were run on a single NVIDIA RTX 4090 24GB GPU.

\subsection{Quantitative Comparison}

\paragraph{Main Results.}
Table~\ref{tab:main-results} reports results on the same benchmark. The main trend is that answer correctness and evidence grounding do not improve together. Across the baselines, the strongest answer, evidence, and combined results come from different model-method combinations: the best baseline answer score is 33.65, the best evidence score is 13.01, and the best combined score is 5.96. This shows that stronger answer prediction does not necessarily imply stronger evidence grounding.

\methodname{} achieves the best performance on answer, evidence, and combined correctness, reaching 38.51 answer, 27.50 evidence, and 14.54 combined score. Compared with the strongest baseline score for each metric, this corresponds to gains of 4.86 points in answer accuracy, 14.49 points in evidence accuracy, and 8.58 points in combined correctness. The substantially larger gains on evidence and combined correctness suggest that the explicit world model mainly helps the system ground its answers in observed visual support. This supports the central claim of \datasetname{}: answer accuracy alone can hide cases where a model predicts a plausible answer but grounds it in the wrong person, timestamp, viewpoint, or box. The conditional Hallucination rate further clarifies this gap: the lowest baseline hallucination rate is 81.22, while \methodname{} lowers it to 62.25. Despite these improvements, the absolute combined score remains low, indicating substantial room for future progress in active evidence-grounded reasoning.

\paragraph{Answer-Evidence Gap.}
The answer-evidence gap is the key empirical signal exposed by \datasetname{}. VLM baselines can achieve moderate answer accuracy while providing much weaker target-grounding evidence: the strongest baseline answer score reaches 33.65, while the strongest baseline evidence score is 13.01 and the strongest combined score is only 5.96. This shows that many answer-correct predictions are not supported by correct visual evidence. \methodname{} reduces this gap by maintaining an observation-grounded world model as memory, improving evidence to 27.50 and combined correctness to 14.54. However, its answer score remains higher than its evidence score, indicating that active evidence grounding is still challenging even with an explicit world model.

\paragraph{Backbone VLM Comparison.}
Table~\ref{tab:main-results} shows that backbone choice affects performance, but does not by itself solve evidence grounding. Across VLM baselines, answer scores differ only moderately, while evidence and combined scores remain substantially lower. Different model-method combinations perform best on different metrics: \texttt{Qwen3-VL 8B} with \baselineReact{} gives the best answer score of 33.65, \texttt{Qwen3-VL 4B} with \baselineReact{} gives the best evidence score of 13.01, and \texttt{Qwen3.5 9B} with \baselineReact{} gives the best combined score of 5.96. The lowest baseline hallucination rate, 81.22, is obtained by \texttt{Qwen3.5 9B} with \baselineSearch{}. This variation across metrics further shows that stronger VLM backbones alone are insufficient for reliable evidence grounding. \datasetname{} itself is method-agnostic and does not assume a VLM-based solution.

\begin{table}[t]
\centering
\caption{
Quantitative comparison on the \datasetname{}. All scores are percentages.
Hallucination rate is the conditional rate of answer-correct predictions without correct evidence,
computed as $(\mathrm{Answer}-\mathrm{Combined})/\mathrm{Answer}$, where lower is better.
}
\label{tab:main-results}
\scalebox{0.9}{\begin{tabular}{cc|ccc|c}
\toprule[0.4mm]
\rowcolor{mygray} Backbone VLM & Method & Answer $\uparrow$ & Evidence $\uparrow$ & Combined $\uparrow$ & Hallucination $\downarrow$ \\
\hline
\multirow{4}{*}{\shortstack{Gemma-4 E2B\\(5B)}} 
& \baselineMonolithic{} & 29.12 & 0.05 & 0.05 & 99.83 \\
& \baselineCot{}        & 25.31 & 0.05 & 0.05 & 99.81 \\
& \baselineSearch{}     & 26.69 & 4.00 & 0.81 & 96.96 \\
& \baselineReact{}      & 29.50 & 5.67 & 1.91 & 93.54 \\
\hline
\multirow{4}{*}{\shortstack{Gemma-4 E4B\\(8B)}}
& \baselineMonolithic{} & 29.50 & 0.14 & 0.14 & 99.52 \\
& \baselineCot{}        & 31.22 & 0.29 & 0.10 & 99.69 \\
& \baselineSearch{}     & 29.27 & 6.10 & 2.29 & 92.18 \\
& \baselineReact{}      & 30.60 & 9.29 & 3.53 & 88.47 \\
\hline
\multirow{4}{*}{Qwen3-VL 4B}
& \baselineMonolithic{} & 25.41 & 2.67 & 1.10 & 95.68 \\
& \baselineCot{}        & 27.74 & 3.19 & 1.48 & 94.67 \\
& \baselineSearch{}     & 21.40 & 11.01 & 3.53 & 83.52 \\
& \baselineReact{}      & 27.74 & 13.01 & 4.29 & 84.54 \\
\hline
\multirow{4}{*}{Qwen3-VL 8B}
& \baselineMonolithic{} & 26.93 & 1.57 & 0.95 & 96.46 \\
& \baselineCot{}        & 27.74 & 1.72 & 0.95 & 96.56 \\
& \baselineSearch{}     & 21.88 & 10.49 & 2.62 & 88.02 \\
& \baselineReact{}      & 33.65 & 11.87 & 4.77 & 85.84 \\
\hline
\multirow{4}{*}{Qwen3.5 4B}
& \baselineMonolithic{} & 27.41 & 4.00 & 2.00 & 92.70 \\
& \baselineCot{}        & 24.59 & 3.86 & 1.72 & 93.02 \\
& \baselineSearch{}     & 24.55 & 9.34 & 3.48 & 85.83 \\
& \baselineReact{}      & 31.22 & 10.49 & 5.24 & 83.21 \\
\hline
\multirow{5}{*}{Qwen3.5 9B}
& \baselineMonolithic{} & 25.88 & 6.29 & 3.05 & 88.21 \\
& \baselineCot{}        & 25.60 & 5.72 & 2.10 & 91.81 \\
& \baselineSearch{}     & 23.36 & 12.11 & 4.39 & 81.22 \\
& \baselineReact{}      & 32.98 & 11.82 & 5.96 & 81.94 \\
& \methodname{} (Ours) & \textbf{38.51} & \textbf{27.50} & \textbf{14.54} & \textbf{62.25} \\
\bottomrule[0.4mm]
\end{tabular}}
\end{table}

\paragraph{Ablation Study.}
Table~\ref{tab:planned-ablations} studies two high-level components of \methodname{}. \emph{Active Observation} denotes active test-time use of the \texttt{observe(sequence, frame\_index, angle\_deg)} interface. \emph{World Model} denotes the explicit observation-grounded world model used by \methodname{}, including graph planning, entities, attributes, and relations. The full agent uses both components. The non-active variant keeps the world model but replaces active observation with a predefined passive observation, and the no-world-model variant keeps active observation but removes the persistent graph planning and world model. The ablation results show that both components improve evidence-grounded reasoning: the world model increases evidence and combined scores, while active observation further improves grounded support. Both components also reduce the conditional hallucination rate, indicating that correct answers are more often supported by valid evidence in the full agent.

\begin{table}[t]
\centering
\caption{
Ablation study of \methodname{}. All scores are percentages.
\emph{Active} denotes active observation.
\emph{World Model} denotes the explicit observation-grounded world model.
Hallucination is as before.
}
\label{tab:planned-ablations}
\begin{tabular}{cc|ccc|c}
\toprule[0.4mm]
\rowcolor{mygray} Active & World Model & Answer $\uparrow$ & Evidence $\uparrow$ & Combined $\uparrow$ & Hallucination $\downarrow$ \\
\hline
$\times$ & $\times$     & 25.31 & 6.86  & 2.72  & 89.25 \\
$\times$ & \checkmark   & 26.83 & 8.06  & 3.62  & 86.51 \\
\checkmark & $\times$   & 37.66 & 18.22 & 9.20  & 75.57 \\
\checkmark & \checkmark & \textbf{38.51} & \textbf{27.50} & \textbf{14.54} & \textbf{62.25} \\
\bottomrule[0.4mm]
\end{tabular}
\end{table}

\paragraph{Failure Analysis.}
The remaining errors highlight the challenge posed by \datasetname{}. Even with an explicit world model, active visual reasoning still requires precise evidence binding across people, frames, viewpoints, and temporal segments. Questions are challenging because they require both selecting the right observations and grounding the final answer in the correct visual support. The results show that \datasetname{} remains a challenging testbed for future work on active visual reasoning.

\section{Conclusion}
\label{sec:conclusion}
\datasetname{} introduces a benchmark for active visual reasoning in real-world embodied environments, where systems must decide what to observe and support their answers with visual evidence. By reporting answer, evidence, combined correctness, and conditional hallucination, \datasetname{} exposes unsupported answers that answer-only evaluation would hide. Our results show that \methodname{}, a world model active reasoning baseline, improves evidence grounding over VLM baselines, while the remaining gap highlights the difficulty of active observation and evidence grounding.

\noindent\textbf{Limitation.}
\datasetname{} evaluates active reasoning over recorded observations rather than live robot control.

\noindent\textbf{Broader Impact.}
We hope \datasetname{} supports future research on embodied active visual reasoning by encouraging methods to ground answers in visual evidence.

\section*{Acknowledgments}

This research was supported by the DARPA Assured Neuro Symbolic Learning and Reasoning (ANSR) program (FA8750-23-2-1016), ONR Global X-Challenge Grant (N62909-25-1-2067), and with the assistance of resources from Monash University and National Computational Infrastructure (NCI Australia) allocation scheme.

\section*{Competing Interests}

We declare that we have no competing interests.

\clearpage

\bibliographystyle{plainnat}
\bibliography{references}

\clearpage

\appendix
\input{supplementary}



\end{document}

%% file: supplementary.tex
\begin{center}
{\LARGE \bfseries Supplementary Material}\par
\end{center}

\noindent This supplementary material provides additional details for \datasetname{}, including question generator definitions, additional benchmark analysis, and baseline execution/prompt templates.

\section{Question Candidate Generator Details}
\label{sec:supp-generators}

The seven VQA generators in \datasetname{} are built from per-sequence scene graphs rather than free-form text generation. Each generated question is paired with the answer format required for evaluation, timestamp metadata, and verifiable target grounding whenever grounding is applicable. Multiple-choice questions use a strict 1-based choice index, while other questions use frame identification, temporal localization, frame-viewpoint pair, or numeric-value outputs. After generation, each candidate is re-validated against the source JRDB annotations, and only questions with recoverable evidence and a well-defined answer format are kept.

\begin{table*}[b]
\centering
\caption{
Generator details for the seven VQA generator families used in \datasetname{}.
Counts refer to the selected 2,098-question benchmark used in the paper.
}
\label{tab:generator-details}
\scriptsize
\setlength{\tabcolsep}{4pt}
\begin{tabular}{>{\raggedright\arraybackslash}p{0.12\linewidth} r >{\raggedright\arraybackslash}p{0.17\linewidth} >{\raggedright\arraybackslash}p{0.16\linewidth} >{\raggedright\arraybackslash}p{0.4\linewidth}}
\toprule
Family & Count & Template(s) & Answer mode & Active reasoning requirement \\
\midrule
Action boundary & 67 & start, span & Frame identification / temporal localization & Localize the onset of an action or recover its full temporal interval instead of answering from a single static frame. \\
Chain 2-hop & 76 & action & Multiple choice & Follow a unique two-hop relational path from the anchor person to the final target before reading out the answer. \\
Chain 3-hop & 222 & action & Multiple choice & Extend the path to three hops, so the answer depends on correctly grounding an additional intermediate person. \\
Chain fork-join & 32 & action & Multiple choice & Ground two branches from the same anchor and merge them at a shared target, requiring branch consistency rather than a single linear chain. \\
Chain hybrid & 51 & action search & Frame-viewpoint pair & Combine a spatial chain with temporal search and viewpoint selection, then return both the decisive frame and the best supporting viewpoint. \\
Long range & 30 & chain, consequence, gap & Multiple choice / numeric value & Recover a target over a larger temporal gap and either predict a later action/consequence or a temporally defined numeric gap. \\
Unique anchor & 1,620 & action change, presence, distance & Multiple choice & Recover a target from a unique identifying anchor and then reason about the same person at another time or state. \\
\bottomrule
\end{tabular}
\end{table*}

\paragraph{Action boundary.}
This family has two evaluated templates. The first asks for the frame at which a target action starts, and the second asks for the full start--end span of that action. Kept questions require stable action segments from the source annotations and reject nearby short breaks, so the benchmark does not reward trivial boundary guesses from noisy or short-lived state changes.

\paragraph{Chain families.}
Chain 2-hop, Chain 3-hop, and Chain fork-join all use multiple-choice action answers, but each requires a distinct relational structure to be grounded before answering. These generators require a unique demographic or relational anchor, a unique annotation-supported path through intermediate people, and non-redundant hops. Thus, the answer must depend on following the intended path rather than shortcutting directly from the anchor to the final target.

\paragraph{Chain hybrid.}
Chain hybrid combines spatial chaining with temporal and viewpoint search. The benchmark first uses a 2-hop path to identify the correct person, then searches over time for the first frame satisfying the queried action condition, and finally requires the viewpoint that best exposes the target. This is the only generator family evaluated with a frame-viewpoint pair answer mode, and all selected Chain hybrid questions are viewpoint-sensitive.

\paragraph{Unique anchor and long range.}
Unique-anchor questions first identify a target from an anchor that remains unique under the available evidence, then ask about the same person's action change, presence, or distance at a later point. Long-range questions instead emphasize temporally distant dependencies, with separate templates for action consequence, chain-style delayed reasoning, and numeric gap prediction.

\section{Additional Results}

\paragraph{Generator Type Analysis.}
Table~\ref{tab:generator-breakdown} groups the generator families into higher-level reasoning types. Action-boundary questions achieve the highest evidence and combined scores for \methodname{}, reaching 56.72 evidence and 23.88 combined score, while the answer score is 37.31. Chain-family questions achieve 41.82 answer, but evidence and combined scores remain much lower at 8.48 and 5.45, reflecting the difficulty of grounding multi-hop relational paths. Chain hybrid is particularly challenging, with 11.76 answer, 5.88 evidence, and zero combined correctness, while long-range questions reach 50.00 answer but only 3.33 evidence and zero combined correctness. Unique-anchor questions achieve 38.52 answer, 31.30 evidence, and 16.73 combined score, indicating substantially stronger grounding than the relational and long-range families. 

\begin{table}[t]
\centering
\caption{
\methodname{} breakdown by high-level generator family.
All scores are percentages.
Hallucination rate is computed as $(\mathrm{Answer}-\mathrm{Combined})/\mathrm{Answer}$, where lower is better.
}
\label{tab:generator-breakdown}
\begin{tabular}{lrrrrr}
\toprule
Generator & Count & Answer $\uparrow$ & Evidence $\uparrow$ & Combined $\uparrow$ & Hallucination $\downarrow$ \\
\midrule
Action boundary & 67    & 37.31 & 56.72 & 23.88 & 36.00 \\
Chain family    & 330   & 41.82 & 8.48  & 5.45  & 86.96 \\
Chain hybrid    & 51    & 11.76 & 5.88  & 0.00  & 100.00 \\
Long range      & 30    & 50.00 & 3.33  & 0.00  & 100.00 \\
Unique anchor   & 1,620 & 38.52 & 31.30 & 16.73 & 56.57 \\
\bottomrule
\end{tabular}
\end{table}
    
\section{Baseline Details}
\label{sec:supp-baselines}

\paragraph{Policies and backbones.}
The reported benchmark compares four baseline methods together with \methodname{}: \baselineMonolithic{}, \baselineCot{}, \baselineSearch{}, and \baselineReact{}. All methods are evaluated on the same 2,098 benchmark questions and share the same answer-evidence scoring protocol. Across baselines, the paper reports six VLM backbones: Gemma-4 E2B (5B), Gemma-4 E4B (8B), Qwen3-VL 4B, Qwen3-VL 8B, Qwen3.5 4B, and Qwen3.5 9B. \baselineMonolithic{} and \baselineCot{} are fixed-context baselines because they do not expose an action interface, so they receive sampled panorama frames as passive visual input. \baselineSearch{} and \baselineReact{} use the public observation interface, with \baselineSearch{} making one proposal-driven observation and \baselineReact{} using an iterative tool-use loop. The following subsections summarize the execution order and main prompt templates used by each baseline.

\paragraph{Shared answer-evidence contract.}
All completed predictions store the routed answer payload together with target-grounding evidence. The routed answer depends on the answer mode of the question: multiple-choice questions return a 1-based choice index, frame-identification questions return a frame id, temporal-localization questions return start and end frame ids, frame-viewpoint-pair questions return a frame id and viewpoint angle, and numeric-value questions return a numeric string. Evidence is scored through the target-grounding output under the evaluation protocol in Section~\ref{sec:benchmark_protocol}, rather than through free-form explanation text.

\subsection{\baselineMonolithic{}}

\paragraph{Execution process.}
The monolithic baseline is a fixed-context single-pass method. It first samples a set of panorama frames and renders them as the only visual context. For each VQA question, it then issues one structured generation step that asks the model to emit the routed answer and exactly one target-person evidence box in the same JSON object. Thus the entire baseline is one fixed-input pass over sampled panorama context, with no active observation or intermediate reasoning stage.

\paragraph{Main prompt.}
The prompt is summarized by its role, visible inputs, and required output before giving the template.

\begin{prompt}{\baselineMonolithic{} final answer stage}
You are answering an active visual reasoning VQA question and localizing the single answer-bearing target person from panorama images.\\
Return exactly one JSON object with the final answer field and one \texttt{evidence\_bboxes} record.\\
Return exactly one bbox for the target person that the question refers to.\\
Frame IDs must be chosen only from the runtime-visible panorama frames listed below.\\
Bboxes must use normalized current-view \texttt{xyxy} coordinates with each integer in \texttt{[0,1000]}.\\
Do not include markdown, prose, wrapper tags, \texttt{track\_id}, or extra keys.\\
Question: \texttt{<question>}\\
Choices:\\
1. \texttt{<choice 1>}\\
2. \texttt{<choice 2>}\\
...\\
Runtime-visible panorama frames: \texttt{<frame\_1, ..., frame\_n>}\\
Ground the target person mentally before answering, then emit the answer and the supporting bbox in the same JSON object.\\
Return the JSON result now.
\end{prompt}

For non-multiple-choice VQA questions, the same prompt body swaps in the answer-mode-specific schema lines. For example, a frame answer uses a JSON shape of the form \texttt{\{"answer":\{"frame\_id":"<supported\_frame\_id>"\}, ...\}}, while temporal localization uses \texttt{start\_frame} and \texttt{end\_frame} keys.

\subsection{\baselineCot{}}

\paragraph{Execution process.}
The Chain-of-Thought baseline reuses the same sampled panorama frames as \baselineMonolithic{}, but inserts an explicit reasoning stage before the final answer-evidence stage. It first asks the model for a short structured reasoning list over the sampled panorama images, and then feeds that reasoning summary into a final answer-plus-bbox stage. The method therefore adds one intermediate reasoning step while remaining a fixed-context baseline with no active observation.

\paragraph{Reasoning prompt.}
The reasoning stage produces a concise structured summary from the sampled panorama context.

\begin{prompt}{\baselineCot{} reasoning stage}
You are producing concise reasoning for an active visual reasoning baseline.\\
Return exactly one JSON object with the shape \texttt{\{"reasoning": ["fact one", "fact two"]\}}.\\
Each step must be short and factual.\\
Do not use step labels, nested objects, or extra keys.\\
When a reasoning step mentions a \texttt{frame\_id}, use only a runtime-visible panorama frame listed below.\\
Question: \texttt{<question>}\\
Choices:\\
1. \texttt{<choice 1>}\\
2. \texttt{<choice 2>}\\
...\\
Runtime-visible panorama frames: \texttt{<frame\_1, ..., frame\_n>}\\
Return the JSON result now.
\end{prompt}

\paragraph{Final answer prompt.}
After reasoning, the final-stage prompt asks the model to produce the routed answer and supporting evidence, conditioned on the reasoning summary.

\begin{prompt}{\baselineCot{} final answer stage}
You are answering an active visual reasoning VQA question and localizing the single answer-bearing target person from panorama images.\\
Return exactly one JSON object with the final answer field and one \texttt{evidence\_bboxes} record.\\
Return exactly one bbox for the target person that the question refers to.\\
Frame IDs must be chosen only from the runtime-visible panorama frames listed below.\\
Bboxes must use normalized current-view \texttt{xyxy} coordinates with each integer in \texttt{[0,1000]}.\\
Do not include markdown, prose, wrapper tags, \texttt{track\_id}, or extra keys.\\
Question: \texttt{<question>}\\
Choices:\\
1. \texttt{<choice 1>}\\
2. \texttt{<choice 2>}\\
...\\
Reasoning summary: \texttt{<fact\_1 | fact\_2 | ...>}\\
Runtime-visible panorama frames: \texttt{<frame\_1, ..., frame\_n>}\\
Ground the target person mentally before answering, then emit the answer and the supporting bbox in the same JSON object.\\
Return the JSON result now.
\end{prompt}

The completed prediction follows the shared answer-evidence contract, while the reasoning summary serves only as an intermediate conditioning artifact for the final stage.

\subsection{\baselineSearch{}}

\paragraph{Execution process.}
The \baselineSearch{} baseline uses three model-generation stages with observation calls. It firstly observes a set of unified sampled frames and predicts one coarse target bbox from that panorama context. The coarse proposal is then converted into one observation through the benchmark observation interface. Given the returned observation view and the original question payload, the model predicts one crop-local refinement bbox. Finally, the model receives the selected refined crop observation, and the question payload, and produces the final answer with the localized target. Thus the execution order is search proposal $\rightarrow$ one observation $\rightarrow$ grounding refinement $\rightarrow$ final answer.

\paragraph{Panorama proposal prompt.}

\noindent The first stage predicts a coarse target-person proposal from the initial visual input.

\begin{prompt}{\baselineSearch{} panorama proposal stage}
You are producing a coarse target-person proposal for an active visual reasoning VQA question from panorama evidence frames.\\
Return exactly one JSON object with the shape \texttt{\{"evidence\_bboxes": [\{"frame\_id": "<frame>", "bbox": [x1, y1, x2, y2]\}]\}}.\\
Return exactly one coarse bbox for the target person referred to by the question.\\
Bboxes must use normalized current-view \texttt{xyxy} coordinates with each integer in \texttt{[0,1000]}.\\
Do not include a reasoning key.\\
Question: \texttt{<question>}\\
Choices:\\
1. \texttt{<choice 1>}\\
2. \texttt{<choice 2>}\\
...\\
Runtime-visible panorama frames: \texttt{<frame\_1, ..., frame\_n>}\\
Return the JSON result now.
\end{prompt}

\paragraph{Crop refinement prompt.}

\noindent The second stage refines the proposal inside the returned crop observation.

\begin{prompt}{\baselineSearch{} crop refinement stage}
You are refining a runtime-produced target-person proposal inside a square crop for an active visual reasoning VQA question.\\
Return exactly one JSON object with the shape \texttt{\{"evidence\_bboxes": [\{"frame\_id": "<frame>", "bbox": [x1, y1, x2, y2]\}]\}}.\\
The bbox must use normalized current-view \texttt{xyxy} coordinates with each integer in \texttt{[0,1000]}.\\
Use only the visible crop observation to refine the proposal. Do not assume any hidden gold target box.\\
Do not include a reasoning key.\\
Question: \texttt{<question>}\\
Choices:\\
1. \texttt{<choice 1>}\\
2. \texttt{<choice 2>}\\
...\\
Crop \texttt{frame\_id}: \texttt{<crop\_frame>}\\
Crop size: \texttt{480}\\
Image 1 is the target crop whose bbox you must refine.\\
Return exactly one bbox in the JSON result.
\end{prompt}

\paragraph{Final answer prompt.}

\noindent The final stage answers the question using the refined target observation.

\begin{prompt}{\baselineSearch{} final answer stage}
You are answering an active visual reasoning VQA question after a runtime proposal has been refined from real observations.\\
Return exactly one JSON object with the routed answer field.\\
Image 1 is the panorama frame selected by the runtime localization proposal.\\
Image 2 is the zoomed crop observation used to refine that same proposal.\\
Runtime-refined target \texttt{frame\_id}: \texttt{<refined\_frame>}\\
Runtime-refined stitched bbox \texttt{xywh}: \texttt{[x, y, w, h]}\\
Runtime-visible panorama frames: \texttt{<frame\_1, ..., frame\_n>}\\
Question: \texttt{<question>}\\
Choices:\\
1. \texttt{<choice 1>}\\
2. \texttt{<choice 2>}\\
...\\
Answer the question about the runtime-refined target person shown in the crop.\\
Do not include markdown, prose, wrapper tags, or extra keys.\\
Return the JSON result now.
\end{prompt}

\subsection{\baselineReact{}}

\paragraph{Execution process.}
The ReAct baseline begins from the question only, without pre-sampled panorama frames or crop observations, and repeatedly alternates between tool-use decisions and public observation calls. It initializes an empty observation list, an empty step history, an observation budget, and the legal set of observable frame ids. At every iteration, it predicts exactly one action JSON object. If the model returns \texttt{observe}, the method issues the public observation call and appends the resulting crop plus metadata to the history. If the model returns \texttt{answer}, the loop terminates and the baseline emits the final answer together with one evidence bbox on the most recent observation image. Thus, unlike the fixed-context baselines and the one-shot \baselineSearch{} workflow, \baselineReact{} exposes an iterative tool-use loop.

\paragraph{Action prompt.}
The recurrent action prompt receives the current observation history, step history, budget state, legal frame ids, and question payload, and returns exactly one JSON action.

\begin{prompt}{\baselineReact{} action stage}
You are an active visual reasoning ReAct agent.\\
Return exactly one JSON object.\\
Observe action schema: \texttt{\{"action": "observe", "frame\_id": "<supported\_frame\_id>", "angle\_deg": 180.0\}}.\\
Tool calls used: \texttt{<n>}\\
Budget limit: \texttt{<B>}\\
Prior observations JSON: \texttt{[...]}\\
Prior steps JSON: \texttt{[...]}\\
Observed \texttt{frame\_id}s so far: \texttt{<...>}\\
Remaining non-duplicate observe budget before answering: \texttt{<...>}\\
Choose the next observe action only from the legal \texttt{frame\_id}s for this sequence and split.\\
Use observe to reduce uncertainty before answering.\\
If remaining budget is 0, you must return an answer action now.\\
Question \texttt{answer\_mode}: \texttt{<answer\_mode>}\\
Question: \texttt{<question>}\\
Choices:\\
1. \texttt{<choice 1>}\\
2. \texttt{<choice 2>}\\
...\\
Answer action schema: \texttt{\{"action": "answer", "answer": 1, "evidence\_bboxes": [\{"frame\_id": "<evidence\_frame>", "bbox": [x1, y1, x2, y2]\}]\}}.\\
Any answer bbox must use normalized current-view \texttt{xyxy} coordinates with each integer in \texttt{[0,1000]}, relative to the most recent observation image only.\\
If no observations exist yet, the next action must be \texttt{observe}.\\
Return the JSON action now.
\end{prompt}

For non-multiple-choice VQA questions, the answer action schema line is replaced with the corresponding structured frame, span, frame-viewpoint, or numeric JSON schema. The prompt also contains answer-mode-specific strategy lines. For example, temporal-localization questions add an instruction that both \texttt{start\_frame} and \texttt{end\_frame} must be grounded before answering.